\documentclass[10pt,twocolumn,letterpaper]{article}

\usepackage{cvpr}
\usepackage{times}
\usepackage{epsfig}
\usepackage{graphicx}
\usepackage{amsmath}
\usepackage{amssymb}
\usepackage{arydshln}

\usepackage{comment}
\usepackage{booktabs}

\newcommand{\argmin}{\operatornamewithlimits{arg\,min}}

\usepackage[pagebackref=true,breaklinks=true,letterpaper=true,colorlinks,bookmarks=false]{hyperref}

\cvprfinalcopy 

\def\cvprPaperID{} 

\ifcvprfinal\fi
\begin{document}

\title{Temporal Action Segmentation from Timestamp Supervision}

\author{Zhe Li, Yazan Abu Farha, Juergen Gall\\
University of Bonn, Germany\\
{\tt\small  s6zeliii@uni-bonn.de, \{abufarha,gall\}@iai.uni-bonn.de}
}

\maketitle

\begin{abstract}
Temporal action segmentation approaches have been very successful recently. However, annotating 
videos with frame-wise labels to train such models is very expensive and time consuming. While 
weakly supervised methods trained using only ordered action lists require less annotation 
effort, the performance is still worse than fully supervised approaches. In this paper, we 
propose to use timestamp supervision for the temporal action segmentation task. Timestamps require a
comparable annotation effort to weakly supervised approaches, and yet provide a more 
supervisory signal. To demonstrate the effectiveness of timestamp supervision, we propose 
an approach to train a segmentation model using only timestamps annotations. Our approach 
uses the model output and the annotated timestamps to generate frame-wise labels by detecting the action changes. We further introduce a confidence loss that forces the predicted probabilities to monotonically decrease as the distance to the timestamps increases. This ensures that all and not only the most distinctive frames of an action are learned during training.  
The evaluation on four datasets shows that models trained with 
timestamps annotations achieve comparable performance to the fully supervised approaches.
\end{abstract}

\vspace{-3mm}
\section{Introduction}

\begin{figure}[t]
   \centering
      \includegraphics[width=.6\columnwidth]{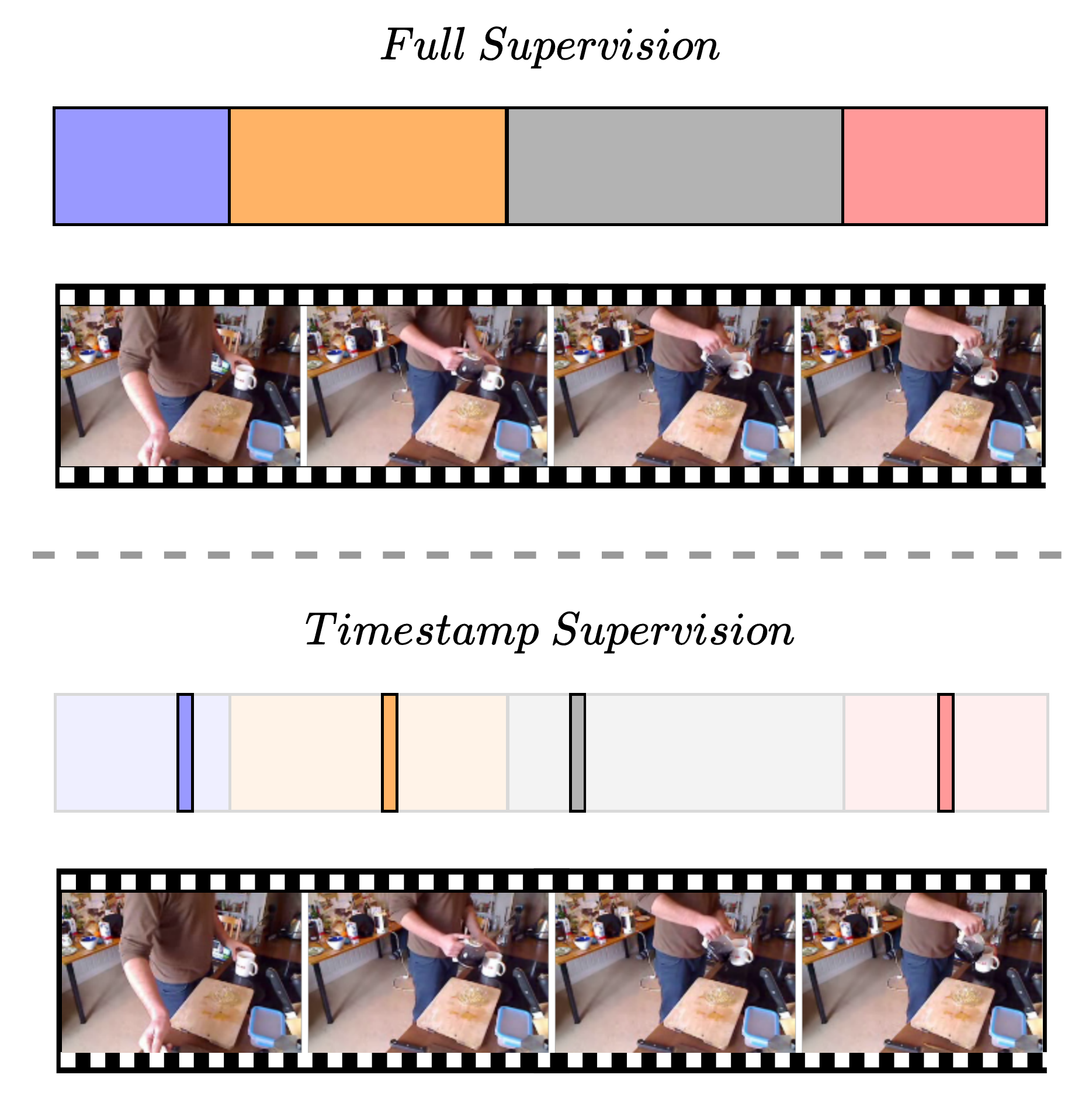}
   \caption{For fully supervised action segmentation, each frame in the training videos 
   is annotated with an action label (top). This process is time-consuming since it requires an accurate annotation of the start and end frame of each action. To reduce the annotation effort, we propose to use timestamps as supervision (bottom). In this case, only one arbitrary frame needs to be annotated for each action and the annotators do not need to search for the start and end frames, which is the most time-consuming annotation part.}
   \label{fig:teaser}
   \vspace{-3.5mm}
\end{figure}

Analyzing 
and understanding video content is very important for many applications, such as surveillance or 
intelligent advertisement. Recently, several approaches have been very successful in analyzing and 
segmenting activities in videos~\cite{lea2017temporal, lei2018temporal, farha2019ms, mac2019learning, wang2020boundary}. 
Despite the success of the previous approaches, they rely on fully annotated videos where the start and 
end frames of each action are annotated. 

This level of supervision, however, is very time consuming and hard to 
obtain. Furthermore, as the boundaries between action segments are usually ambiguous, this might result 
in inconsistencies between annotations obtained from different annotators. To alleviate these problems, 
many researchers start exploring weaker levels of supervision in the form of 
transcripts~\cite{bojanowski2014weakly, richard2018neuralnetwork, li2019weakly} or even sets~\cite{richard2018action, fayyaz2020set, li2020set}. For transcript-level supervision, 
the videos are annotated with an ordered list of actions occurring in the video without the starting 
and ending time of each action. Whereas for the set-level supervision, only the set of actions are 
provided without any information regarding the order or how many times each action occurs in the videos.

While transcript-level and set-level supervision significantly reduce the annotation effort, the 
performance is not satisfying and there is still a gap compared to fully supervised approaches. 
In this paper, inspired by the recently introduced timestamp supervision for action recognition~\cite{moltisanti2019action}, we propose to use timestamp supervision for the action segmentation task to address the limitations of the current weakly supervised approaches. For timestamp supervision, 
only one frame is annotated from each action segment as illustrated in Fig.~\ref{fig:teaser}. 
Such timestamps annotations can be obtained with comparable effort to transcripts, 
and yet it provides more supervision. Besides the ordered list of actions occurring in the video, timestamps 
annotations give partial information about the location of the action segments, which can be 
utilized to further improve the performance. 

Given the timestamps annotations, the question is how to train a segmentation model with such level of supervision. A naive approach takes only the sparsely annotated frames for training. This, however, ignores most of the information in the video and does not achieve good results as we will show in the experiments. Another strategy is to iterate the process and consider frames with high confidence scores near the annotations as additional annotated frames and include them during training~\cite{moltisanti2019action}. Furthermore, frames that are far away from the annotations can be considered as negative samples~\cite{ma2020sf}. For temporal action segmentation, which is comparable to semantic image segmentation, however, all frames need to be annotated and there are no large parts of the video that can be used to sample negative examples. Furthermore, relying only on frames with high confidence discards many of the video frames that occur during an action and focuses only on the most distinctive frames of an action, which can be sufficient for action recognition or detection but not for action segmentation.  

In this work, we therefore propose a different approach where all frames of the videos are used. Instead of detecting frames of high confidences, we aim to identify changes of actions in order to divide the videos into segments. Since for each action change the frames before the change should be assigned to the previous timestamp and after the change to the next timestamp, we find the action changes by minimizing the variations of the features within each of the 
two clusters of frames. While we can then train the model on all frames by assigning the label of the timestamp to the corresponding frames, it does not guarantee that all frames of an action are effectively used. We therefore introduce a loss function that enforces a monotonic decrease in the class probabilities as the distance to the timestamps increases. This loss encourages the model to predict higher probabilities for low confident regions that are surrounded by high confident frames and therefore to use all frames and not only the most distinctive frames.

Our contribution is thus three folded.
\begin{enumerate}
  \item We propose to use timestamp supervision for the temporal action segmentation task, where the goal is to 
  predict frame-wise action labels for untrimmed videos. 
  \item We introduce an approach to train a temporal action segmentation model from timestamp supervision. 
  The approach uses the model predictions and the annotated timestamps for estimating action changes. 
  \item We propose a novel confidence loss that forces the model confidence to decrease monotonically as the distance to the timestamp increases.
\end{enumerate}

We evaluate our approach on four datasets: 50Salads~\cite{stein2013combining}, Breakfast~\cite{kuehne2014language}, BEOID~\cite{damen2014you}, and Georgia Tech Egocentric Activities (GTEA)~\cite{fathi2011learning}. We show that training an action 
segmentation model is feasible with only timestamp supervision without compromising the performance compared to the fully supervised approaches. On the 50Salads dataset, for instance, we achieve 97\% of the accuracy compared to fully supervised learning, but at a tiny fraction of the annotation costs. \footnote{The source code for our model and the timestamps annotations are publicly 
available at \url{https://github.com/ZheLi2020/TimestampActionSeg}.}


\section{Related Work}
We briefly discuss the related work for the temporal action segmentation task 
at different levels of supervision.
\vspace{-.4mm}
\paragraph{Fully Supervised Action Segmentation.}
Temporal action segmentation has received an increasing interest recently. In contrast to action recognition 
where the goal is to classify trimmed videos~\cite{simonyan2014two, carreira2017quo, feichtenhofer2019slowfast}, 
temporal action segmentation requires capturing long-range dependencies to classify each frame in the input video. 
To achieve this goal, many approaches combined frame-wise classifiers with grammars~\cite{vo2014stochastic, pirsiavash2014parsing} 
or with hidden Markov models (HMMs)~\cite{lea2016segmental, kuehne2016end, kuehne2020hybrid}. Despite the success of these 
approaches, their performance was limited and they were slow both at training and inference time. Recent approaches, therefore, 
utilized temporal convolutional networks to capture long-range dependencies for the temporal action segmentation 
task~\cite{lea2017temporal, lei2018temporal}. While such approaches managed to generate accurate predictions, they suffer 
from an over-segmentation problem. To alleviate this problem, current state-of-the-art methods follow a multi-stage 
architecture with dilated temporal convolutions~\cite{farha2019ms, wang2020boundary, li2020ms, ishikawa2020alleviating, huang2020improving}. 
These approaches rely on fully annotated datasets that are expensive to obtain. On the contrary, we 
address the temporal action segmentation task in a weakly supervised setup.

\paragraph{Weakly Supervised Action Segmentation.}
Weakly supervised action segmentation has been an active research area recently. Earlier approaches 
apply discriminative clustering to detect actions using movie scripts~\cite{bojanowski2014weakly, duchenne2009automatic}. 
However, these approaches ignored the action ordering information and only focused on distinguishing action 
segments from background, which is a common practice for the temporal action localization task~\cite{lee2020background, lee2021weakly}. Bojanowski~\etal~\cite{bojanowski2014weakly} extended these ideas to segment 
actions in videos using transcripts in the form of an ordered list of actions as supervision. Recently, many researchers 
addressed the task by aligning the video frames and the transcripts using connectionist temporal 
classification~\cite{huang2016connectionist}, dynamic time warping~\cite{chang2019d3tw}, or energy-based 
learning~\cite{li2019weakly}. 
Other approaches generate pseudo ground truth labels for the training videos and iteratively refine 
them~\cite{kuehne2017weakly, richard2017weakly, ding2018weakly, kuehne2020hybrid}. In~\cite{richard2018neuralnetwork}, 
a frame-wise loss function is combined with the Viterbi algorithm to generate the target labels. While these 
approaches have been very successful, they suffer from a slow inference time as they iterate 
over all the training transcripts and select the one with the highest score. Souri~\etal~\cite{souri2019fast} 
addressed this issue by predicting the transcript besides the frame-wise scores at inference time. 
While these approaches rely on a cheap transcript supervision, their performance is much worse than fully supervised 
approaches. In contrast to these approaches, we propose a higher level of supervision in the form of timestamps 
that can be obtained with comparable effort to the transcript supervision, and yet reduces the gap to the fully 
supervised approaches. 
There is another line of research addressing the action segmentation task from set-level 
supervision~\cite{richard2018action, fayyaz2020set, li2020set}. These approaches opt for a weaker level 
of supervision at the cost of performance. In contrast to these approaches, we propose a good compromise 
between supervision level and performance.

\paragraph{Timestamp Supervision for Recognizing Activities.}
Timestamp supervision has not yet received much attention from the action recognition community. 
Initial attempts were inspired by the success of point supervision for semantic segmentation~\cite{bearman2016s}. 
Mettes~\etal~\cite{mettes2016spot} apply multiple instance learning for spatio-temporal action 
localization using points annotation on a sparse subset of frames. Ch{\'e}ron~\etal~\cite{cheron2018flexible} 
use discriminative clustering to integrate different types of supervision for the spatio-temporal 
action localization task. Recently, Moltisanti~\etal~\cite{moltisanti2019action} proposed a sampling 
distribution based on a plateau function centered around temporal timestamps annotations to train 
a fine-grained action classifier. This approach relies on the classifier response to sample frames 
around the annotated timestamps and uses them for training. The method was tested for classifying 
trimmed videos and also showed promising results for temporal action localization. Ma~\etal~\cite{ma2020sf} 
extended the action localization setup by mining action frames and background 
frames for training.


\section{Temporal Action Segmentation}
Temporal action segmentation is the task of predicting frame-wise action labels for a given 
input video. Formally, given a sequence of video frames $X=[x_1, \dots ,x_T]$, where $T$ 
is the number of frames, the goal is to predict a sequence of frame-wise action labels 
$[a_1, \dots ,a_T]$. In contrast to the fully supervised approaches, 
which assume that the frame-wise labels are given at training time, we consider a weaker level of supervision in the 
form of timestamps. In Section~\ref{sec:timestamps_supervision} we introduce the timestamp supervision 
for the temporal action segmentation task. Then, we describe the proposed framework for learning 
from timestamp supervision in Section~\ref{sec:method}. Finally, we provide the details of the 
loss function in Section~\ref{sec:loss}.


\subsection{Timestamp Supervision}
\label{sec:timestamps_supervision}
In a fully supervised setup, the frame-wise labels \mbox{$[a_1, \dots ,a_T]$} of the training 
videos are available. On the contrary, for timestamp supervision, only a single frame for each 
action segment is annotated. Given a training video $X$ with $T$ frames and $N$ action 
segments, where $N << T$, only $N$ frames are annotated with labels \mbox{$A_{TS}=[a_{t_1}, \dots ,a_{t_N}]$}, 
where frame $t_i$ belongs to the $i\mbox{-}th$ action segment. 
To annotate timestamps, one can go fast forward through a video and press a button 
when an action occurs. This does not take more time than annotating transcripts. Whereas 
annotating the start and end frames of each action requires going slowly back and forth 
between the frames. As reported in~\cite{ma2020sf}, annotators need 6 times longer to annotate 
the start and end frame compared to annotating a single timestamp.
While timestamps are much easier 
to obtain compared to the full annotation of the video frames, they provide much more information 
compared to weaker forms of supervision such as transcripts. 
Fig.~\ref{fig:teaser} illustrates the difference between timestamp supervision and full supervision.

\begin{figure}[tb]
   \centering
      \includegraphics[width=.6\columnwidth]{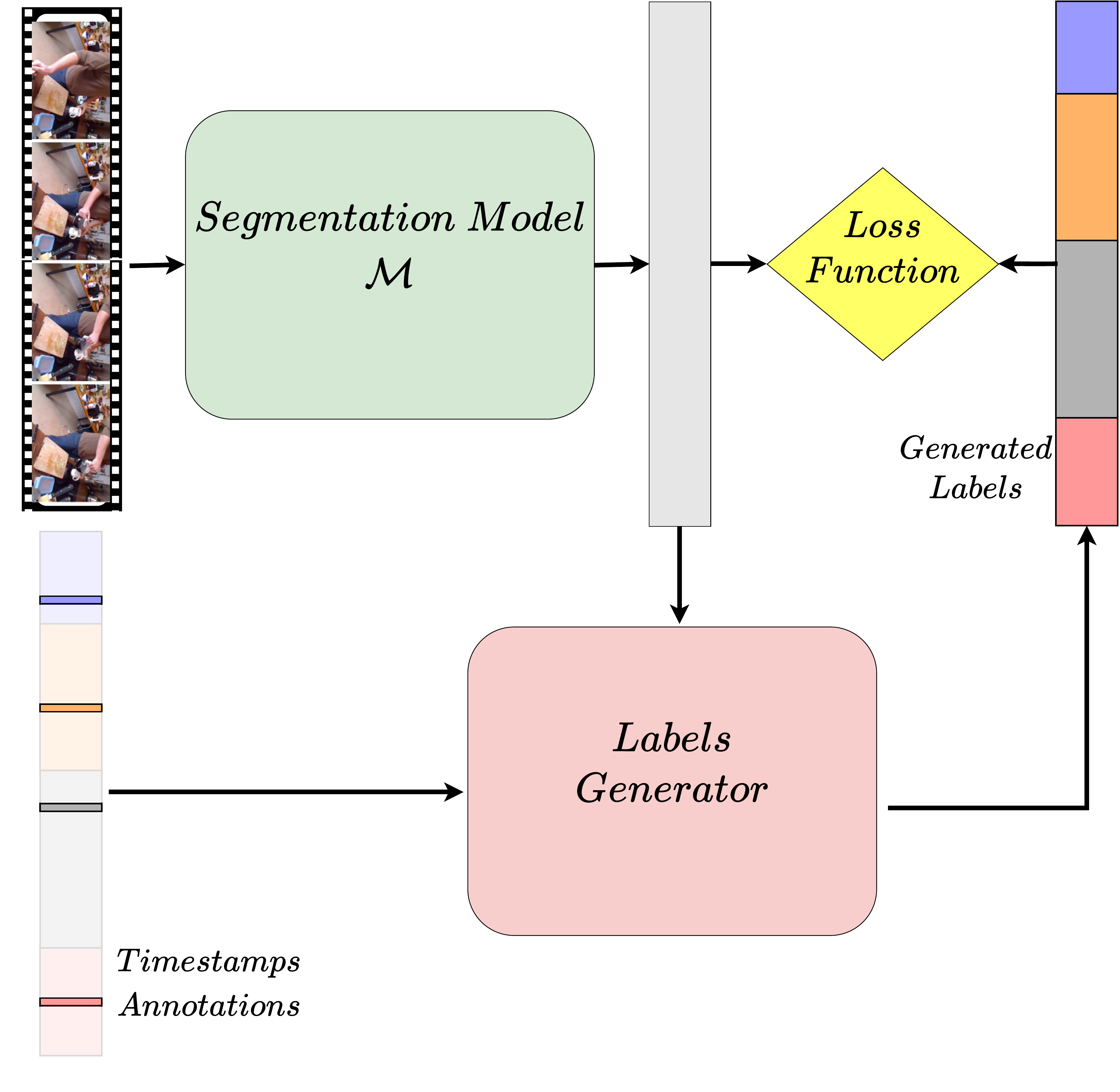}
   \caption{The framework of the proposed approach for training with timestamp supervision. 
   Given the output of the segmentation model and the timestamps annotations, we generate action 
   labels for each frame in the input video by estimating where the action labels change. A loss function is then computed between the 
   predictions and the generated labels.}
   \label{fig:Framework_new}
   \vspace{-4mm}
\end{figure}


\subsection{Action Segmentation from Timestamp Supervision}
\label{sec:method}
Given an action segmentation model $\cal{M}$ and a set of training videos with timestamps annotations, 
the goal is to train the model $\cal{M}$ to predict action labels for each frame in the input video. 
If the frame-wise labels are available during the training, as in the fully supervised case, then it is 
possible to apply a classification loss on the output of the model $\cal{M}$ for each frame in the input 
video. However, in timestamp supervision, only a sparse set of frames are annotated. To alleviate this 
problem, we propose to generate frame-wise labels for the training videos, and use them as a target 
for the loss function as illustrated in Fig.~\ref{fig:Framework_new}.

\begin{figure}[tb]
   \centering
      \includegraphics[trim={0cm 0cm 0cm 1.0cm},clip,width=.9\columnwidth]{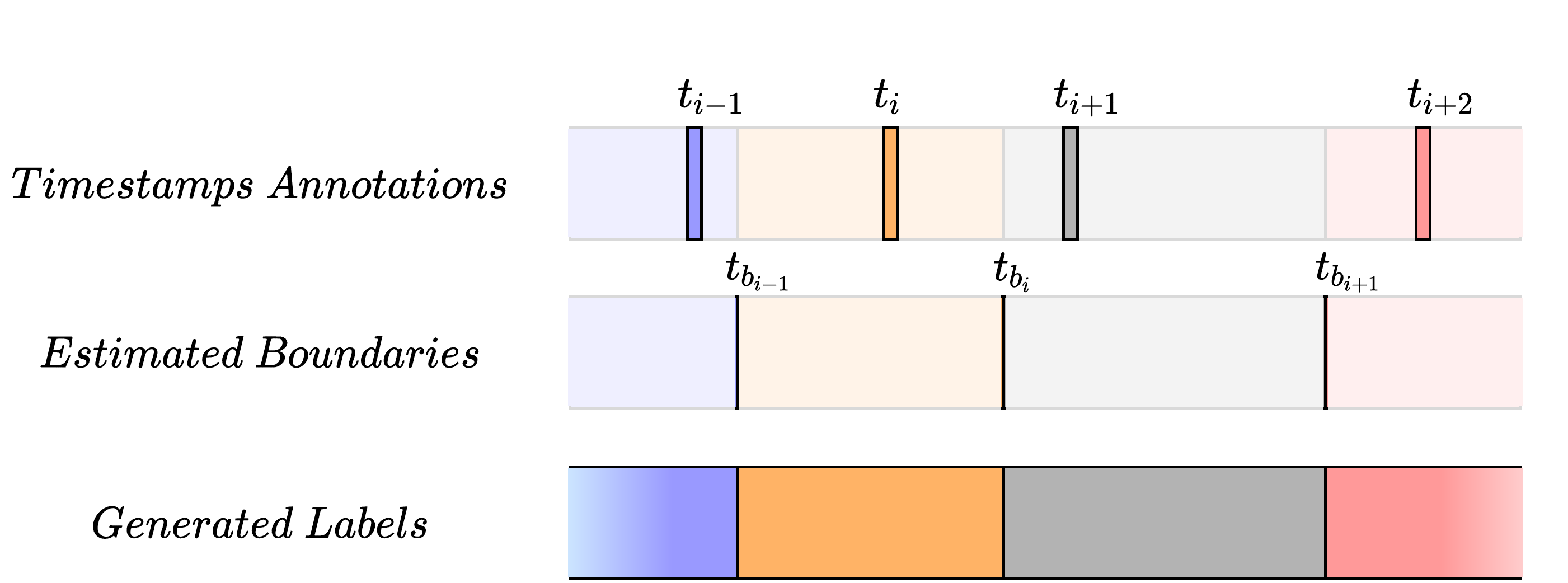}
   \caption{Given the timestamps annotations, we first estimate where the actions change between consecutive 
   timestamps. To generate the frame-wise labels, all the frames that lie between an annotated 
   timestamp and an estimated time of action change are assigned with the same action label as the annotated timestamp.}
   \label{fig:pseudo_gt}
\end{figure}

\paragraph{Detecting Action Changes.} 
Given the timestamps annotations 
\mbox{$A_{TS}=[a_{t_1}, \dots ,a_{t_N}]$} for a video $X$, we want to generate frame-wise action 
labels \mbox{$\hat{A}=[\hat{a}_{1}, \dots ,\hat{a}_{T}]$} for each frame in that video such 
that $\hat{a}_{t_i} = a_{t_i}$ for $i \in [1, N]$. As for each action segment there is an annotated 
frame, finding the frame-wise labels can be reduced to finding the action change between each 
consecutive annotated timestamps. To this end, we pass the input video $X$ to the segmentation 
model $\cal{M}$, which will be described in Section~\ref{sec:implementation_details}, and use the output of the penultimate layer $H$ combined with 
the timestamps annotations to estimate where the action labels change between the timestamps. 
To generate the labels, all the frames that lie between an annotated timestamp and an 
estimated time of action change are assigned with the same action label as the annotated timestamp as illustrated 
in Fig.~\ref{fig:pseudo_gt}. To detect the action change between two timestamps $t_{i}$ and $t_{i+1}$, 
we find the time $t_{b_i}$ that minimizes the following stamp-to-stamp energy function 
\begin{equation} \label{eqn:s2s_energy}
\begin{split}
& t_{b_i} = \argmin_{\hat{t}} \sum_{t=t_i}^{\hat{t}} d(h_{t}, c_i) + \sum_{t=\hat{t}+1}^{t_{i+1}} d(h_{t}, c_{i+1}), \\
s.t. & \\
& c_i = \frac{1}{\hat{t} - t_i + 1} \sum_{t=t_i}^{\hat{t}} h_{t}, \\ 
& c_{i+1} = \frac{1}{t_{i+1} - \hat{t}} \sum_{t=\hat{t}+1}^{t_{i+1}} h_{t}, \\
& t_i \leq \hat{t} < t_{i+1},
\end{split}
\end{equation}
where $d(.,.)$ is the Euclidean distance, $h_{t}$ 
is the output of the penultimate layer at 
time $t$, $c_i$ is the average of the output between the first timestamp $t_i$ and the estimate $\hat{t}$, and 
$c_{i+1}$ is the average of the output between the estimate $\hat{t}$ and the second timestamp $t_{i+1}$. 
\Ie, we find the time $t_{b_i}$ that divides the frames between two timestamps into two clusters with the minimum 
distance between frames and the corresponding cluster center.

\paragraph{Forward-Backward Action Change Detection.} 
In (\ref{eqn:s2s_energy}), the stamp-to-stamp energy function considers only the frames 
between the annotated timestamps to estimate where the actions change. Nonetheless, if we already have an estimate 
for $t_{b_{i-1}}$, then we already know that frames between the estimate $t_{b_{i-1}}$ and the next timestamp $t_{i}$ will be assigned to action label $a_{t_{i}}$. This 
information can be used to estimate the time of action change for the next action segment $t_{b_{i}}$. The same 
argument also holds if we start estimating the boundaries in reverse order. \Ie, if we already 
know $t_{b_{i+1}}$, then frames between $t_{i+1}$ and $t_{b_{i+1}}$ can be used to estimate $t_{b_{i}}$. 
We call the former estimate a forward estimate for the $i\mbox{-}th$ action change, whereas the later is called 
the backward estimate. The final estimate for $t_{b_{i}}$ is the average of these two estimates. 
Formally
\begin{equation} \label{eqn:forward_backward}
\begin{split}
& t_{b_i} = \frac{t_{b_i, FW} + t_{b_i, BW}}{2} \\
s.t. &\\
& t_{b_i, FW} = \argmin_{\hat{t}} \sum_{t=t_{b_{i-1}}}^{\hat{t}} d(h_{t}, c_i) + \sum_{t=\hat{t}+1}^{t_{i+1}} d(h_{t}, c_{i+1}), \\
& t_{b_i, BW} = \argmin_{\hat{t}} \sum_{t=t_i}^{\hat{t}} d(h_{t}, c_i) + \sum_{t=\hat{t}+1}^{t_{b_{i+1}}} d(h_{t}, c_{i+1}). \\
\end{split}
\end{equation}

\subsection{Loss Function}
\label{sec:loss}
Recent fully supervised approaches for action segmentation use a combination of a classification loss 
and a smoothing loss~\cite{farha2019ms, ishikawa2020alleviating, wang2020boundary}. Besides these losses, 
we further introduce a novel confidence loss for the timestamp supervision. In the following, we describe 
in detail each loss function.

\textbf{Classification Loss.} We use a cross entropy loss between the predicted action probabilities 
and the corresponding generated target label
\begin{equation}\label{cross entropy}
    \mathcal{L}_{cls} = \frac{1}{T} \sum_t{-log(\tilde{y}_{t,\hat{a}})},
\end{equation}
where $\tilde{y}_{t,\hat{a}}$ is the predicted probability for the target label $\hat{a}$ at time $t$.

\textbf{Smoothing Loss.} As the classification loss treats each frame independently, it 
might result in an undesired over-segmentation effect. To encourage a smooth transition between 
frames and reduce over-segmentation errors, we use the truncated mean squared error~\cite{farha2019ms} as  a smoothing loss 
\begin{align}
\begin{split}\label{smoothingloss}
    & \mathcal{L}_{T-MSE} = \frac{1}{TC} \sum_{t,a} \Tilde{\Delta}_{t,a}^2,
\end{split}\\
\begin{split}\label{trancated}
    & \Tilde{\Delta}_{t,a}=\left\{
         \begin{array}{ll}
             \Delta_{t,a} &: \Delta_{t,a} \leq \tau  \\
             \tau &: otherwise \\
         \end{array}
\right.
\end{split},\\
    & \Delta_{t,a}=|\log \tilde{y}_{t,a} - \ \log \tilde{y}_{t-1,a}|,
    \label{logdifferences}
\end{align}
where $T$ is the video length, $C$ is the number of action classes, and $\tilde{y}_{t,a}$ 
is the probability of action $a$ at time $t$.

\begin{figure}[tb]
   \centering
      \includegraphics[trim={1.4cm 0cm 0cm 0cm},clip,width=.74\columnwidth]{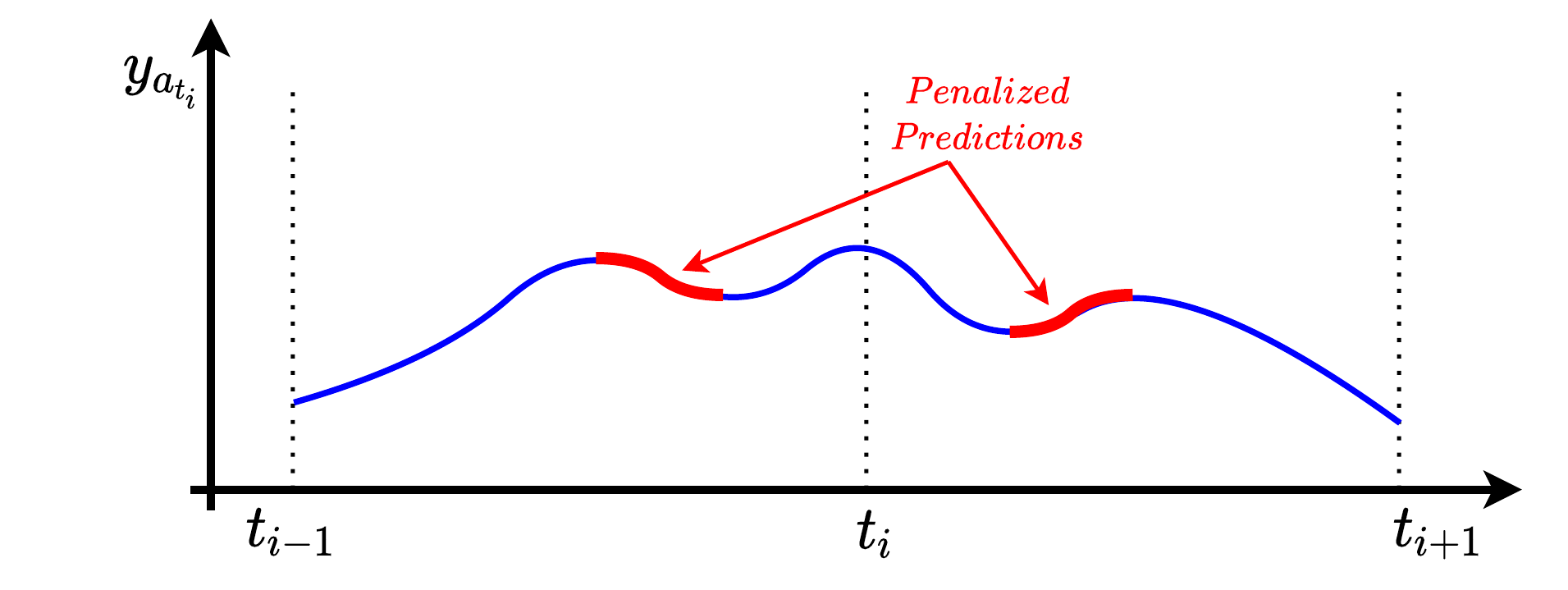}
   \caption{The confidence loss penalizes increases in the model confidence for label $a_{t_i}$ as we 
   move away from the annotated timestamp $t_i$.}
   \label{fig:confidence_loss}
   \vspace{-3mm}
\end{figure}

\textbf{Confidence Loss.} Our approach relies on the model output to detect action changes. Nonetheless, as some frames are more informative than others, the model confidence might alternate between high and low values within the same action segment. Such behavior might result in ignoring regions with low confidence within the segments. To alleviate this problem, we apply the following loss 
\begin{align}
\begin{split}\label{rankingloss}
    & \mathcal{L}_{conf} = \frac{1}{T'} \sum_{\substack{a_{t_i} \in A_{TS}}}{\left(
                                             \sum_{t=t_{i-1}}^{t_{i+1}}{
                                                  \delta_{a_{t_i},t }
                                                  }\right)
                                             },
\end{split}\\
\begin{split}\label{trancatedleft}
    \delta_{a_{t_i}, t} &=\left\{
         \begin{array}{ll}
             max(0,\ \log \tilde{y}_{t,a_{t_i}} - \ \log \tilde{y}_{t-1,a_{t_i}}) &if\ t \ge t_i  \\
             max(0,\ \log \tilde{y}_{t-1,a_{t_i}} - \ \log \tilde{y}_{t,a_{t_i}}) &if\ t < t_i  \\
         \end{array}
\right.
\end{split},
\end{align}
where $\tilde{y}_{t,a_{t_i}}$ is the probability of action $a_{t_i}$ at time $t$, and $T' = 2(t_N-t_1)$ is the 
number of frames that contributed to the loss. For the first and last timestamps, we set $t_0 = t_1$ 
and $t_{N+1} = t_{N}$. This loss penalizes an increase in confidence as we 
deviate from the annotations as illustrated in Fig.~\ref{fig:confidence_loss}. 

Enforcing monotonicity on the model confidence has two effects as shown in Fig.~\ref{fig:effect_of_conf_loss}. First, it encourages the model 
to predict higher probabilities for low confident regions that are surrounded by regions with high confidence. 
Second, it suppresses outlier frames with high confidence that are far from the timestamps and not supported by high 
confident regions.

The final loss function to train the segmentation model is the sum of these three losses

\begin{equation}\label{totalloss}
    \mathcal{L}_{total} = \mathcal{L}_{cls} + \alpha \mathcal{L}_{T-MSE} + \beta \mathcal{L}_{conf},
\end{equation}
where $\alpha$ and $\beta$ are hyper-parameters to balance the contribution of each loss.

\section{Experiments}

\subsection{Datasets and Metrics}
\textbf{Datasets.} We evaluate our approach on four datasets: 50Salads~\cite{stein2013combining}, 
Breakfast~\cite{kuehne2014language}, BEOID~\cite{damen2014you}, and Georgia Tech Egocentric Activities (GTEA)~\cite{fathi2011learning}.

The \textbf{50Salads} dataset contains 50 videos with roughly 0.6M frames, where the frames are annotated with 17 action 
classes. The videos show actors preparing different kind of salads. We use five-fold cross validation 
for evaluation and report the average.

The \textbf{Breakfast} dataset contains 1712 videos with roughly 3.6M frames, where the frames are annotated with 48 action 
classes. All actions are related to breakfast preparation activities. We use the standard four 
splits for evaluation and report the average.

The \textbf{BEOID} dataset contains 58 videos, where the frames are annotated with 34 actions 
classes. For evaluation, we use the same training-testing split as in~\cite{ma2020sf}.

The \textbf{GTEA} dataset contains 28 videos with roughly 32K frames, where the frames are annotated with 11 action 
classes. 

For evaluation, we 
report the average of four splits.

To generate the timestamps annotations, we randomly select one frame from each action 
segment in the training videos. 
We further evaluate our approach using human and noisy annotations in Section~\ref{sec:comparison_with_sota}. Additional settings are evaluated in the supplementary material.

\textbf{Metrics.} We use the standard metrics for fully supervised action segmentation and 
report frame-wise accuracy (Acc), segmental edit distance (Edit) and segmental F1 scores 
at overlapping thresholds $10\%,\ 25\%$ and $50\%$.

\textbf{Baselines.} We implement two baselines: a Naive and a Uniform baseline. 
The \textbf{Naive} baseline computes the loss at the annotated timestamps only and does not 
generate frame-wise labels. Whereas the \textbf{Uniform} baseline generates the frame-wise labels 
by assuming that action labels change at the center frame between consecutive timestamps.


\subsection{Implementation Details}
\label{sec:implementation_details}
We use a multi-stage temporal convolutional network~\cite{farha2019ms} as a segmentation 
model $\mathcal{M}$. Following~\cite{vats2020event}, we use two parallel stages for the first stage 
with kernel size $5$ and $3$ and pass the sum of the outputs to next stages. 
We train our model for 50 epochs with Adam optimizer. To minimize the impact of initialization, 
only the annotated timestamps are used for the classification loss in the first $30$ epochs, and the 
generated labels are used afterwards. The learning rate is set to $0.0005$ and 
the batch size is $8$. For the loss function, we set $\tau=4$, $\alpha=0.15$ as in~\cite{farha2019ms} 
and set $\beta=0.075$. As input for our model, we use the same I3D~\cite{carreira2017quo} features that 
were used in~\cite{farha2019ms}.


\begin{figure*}[tb]
\begin{center}
\begin{tabular}{c}
   \includegraphics[width=.74\linewidth]{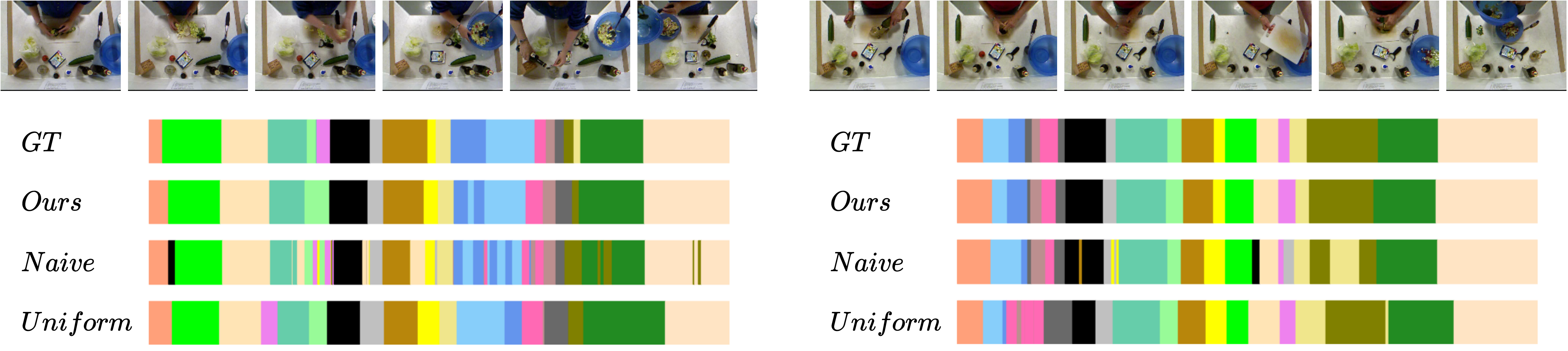}
  	\\
  	(a) 
  	\\
  	\\
  	\includegraphics[width=.74\linewidth]{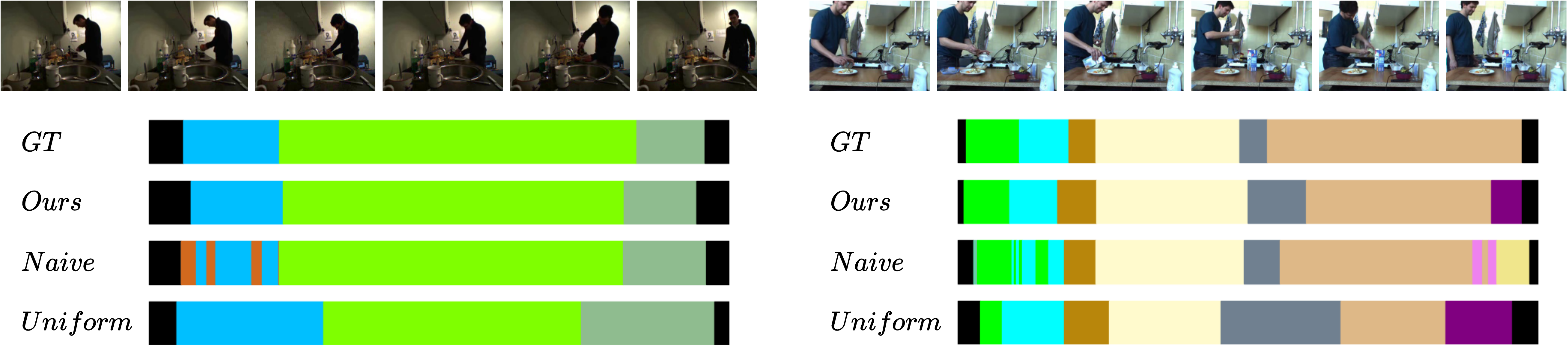}
  	\\
  	(b) 
  	\\
  	\\
  	\includegraphics[width=.74\linewidth]{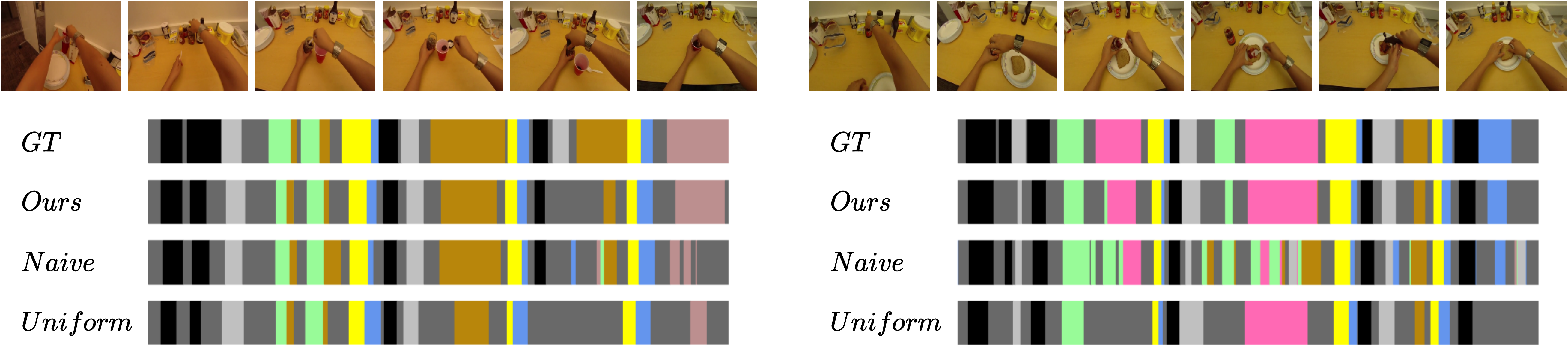}
  	\\
  	(c) 
\end{tabular}
\end{center}
\vspace{-4mm}
   \caption{Qualitative results on (a) 50Salads, (b) Breakfast, and (c) GTEA datasets. As the naive baseline only trains on the sparse annotations, it suffers from an over-segmentation problem. While the uniform baseline reduces this problem by uniformly assigning labels to the frames, the durations of the predicted segments are not accurate and the predictions tend towards a uniform segmentation of the videos. On the contrary, our approach generates better predictions by utilizing the model output to detect where the action labels change.}
\label{fig:qualitative_res}
\end{figure*}

\subsection{Comparison with the Baselines}

In this section, we compare the proposed approach for action segmentation from timestamp 
supervision with the naive and uniform baselines. The results on the three datasets are 
shown in Table~\ref{tab:baselines}. Our approach outperforms these baselines with a large 
margin in all the evaluation metrics. While the naive baseline achieves a good frame-wise 
accuracy, it suffers from a severe over-segmentation problem as indicated by the low F1 and 
Edit scores. This is because it only uses the sparse timestamps annotations for training, which 
leaves a lot of ambiguity for frames without annotations. Using the uniform baseline reduces 
the over-segmentation by uniformly assigning a label for each frame. However, this results 
in inferior frame-wise accuracy as the uniform assignment generates many wrong labels. On the 
contrary, our approach utilizes the model predictions to generate much better target labels, which 
is reflected in the performance as illustrated in Fig.~\ref{fig:qualitative_res}. We also compare the 
performance of our approach to the fully supervised setup in Table~\ref{tab:baselines}. Our 
approach achieves comparable performance to the fully supervised case.

\begin{table}[tb]\setlength{\tabcolsep}{8pt}
   \centering
   \resizebox{.9\columnwidth}{!}{%
      \begin{tabular}{@{\hskip .2in}l@{\hskip .7in}ccc@{\hskip .2in}c@{\hskip .2in}c}
         \toprule
          & \multicolumn{3}{c}{F1@\{10, 25, 50\}} & Edit & Acc  \\
         \midrule
         \emph{\textbf{50Salads}} \\
         \midrule
         Naive & 47.9 & 43.3 & 34.0 & 37.2 & 69.6 \\
         Uniform & 62.9 & 58.2 & 42.3 & 60.4 & 63.4  \\
         Ours &\textbf{73.9} & \textbf{70.9} & \textbf{60.1} & \textbf{66.8} & \textbf{75.6}  \\
         \hdashline
         Full Supervision & 70.8 & 67.7 & 58.6 & 63.8 & 77.8  \\
         \midrule
          \emph{\textbf{Breakfast}} \\
         \midrule
         Naive & 34.1 & 29.1 & 20.1 & 37.4 & 56.8 \\
         Uniform & 66.2 & 56.3 & 36.4 & 68.1 & 51.0 \\
         Ours & \textbf{70.5} & \textbf{63.6} & \textbf{47.4} & \textbf{69.9} & \textbf{64.1}  \\
         \hdashline
         Full Supervision & 69.9 & 64.2 & 51.5 & 69.4 & 68.0  \\
         \midrule
          \emph{\textbf{GTEA}} \\
         \midrule
         Naive & 59.7 & 55.3 & 39.6 & 51.1 & 56.5 \\
         Uniform &  \textbf{78.9} & 72.5 & 50.9 & \textbf{73.1} & 56.5\\
         Ours & \textbf{78.9} & \textbf{73.0} & \textbf{55.4} & 72.3 & \textbf{66.4}  \\
         \hdashline
         Full Supervision & 85.1 & 82.7 & 69.6 & 79.6 & 76.1 \\
         \bottomrule
      \end{tabular}
    }
   \caption{Comparison with the baselines on the three datasets.}
   \vspace{-4mm}
   \label{tab:baselines}
\end{table}


\subsection{Impact of the Loss Function}

The loss function to train our model consists of three losses: a classification loss, a smoothing 
loss, and a confidence loss. Table~\ref{tab:losses} shows the impact of each loss on both the 50Salads 
and the Breakfast dataset. While either of the smoothing loss and the confidence loss gives 
an additional boost in performance, the best performance is achieved when both of the losses are 
combined with the classification loss with a frame-wise accuracy improvement of $2.8\%$ and 
$3.9\%$ on 50Salads and the Breakfast dataset respectively, and roughly $10\%$ on the F1 score 
at $50\%$ overlapping threshold.

\begin{figure}[tb]
   \centering
   \begin{tabular}{c}
         \includegraphics[trim={0.8cm 0cm 0cm 0cm},clip,width=.7\columnwidth]{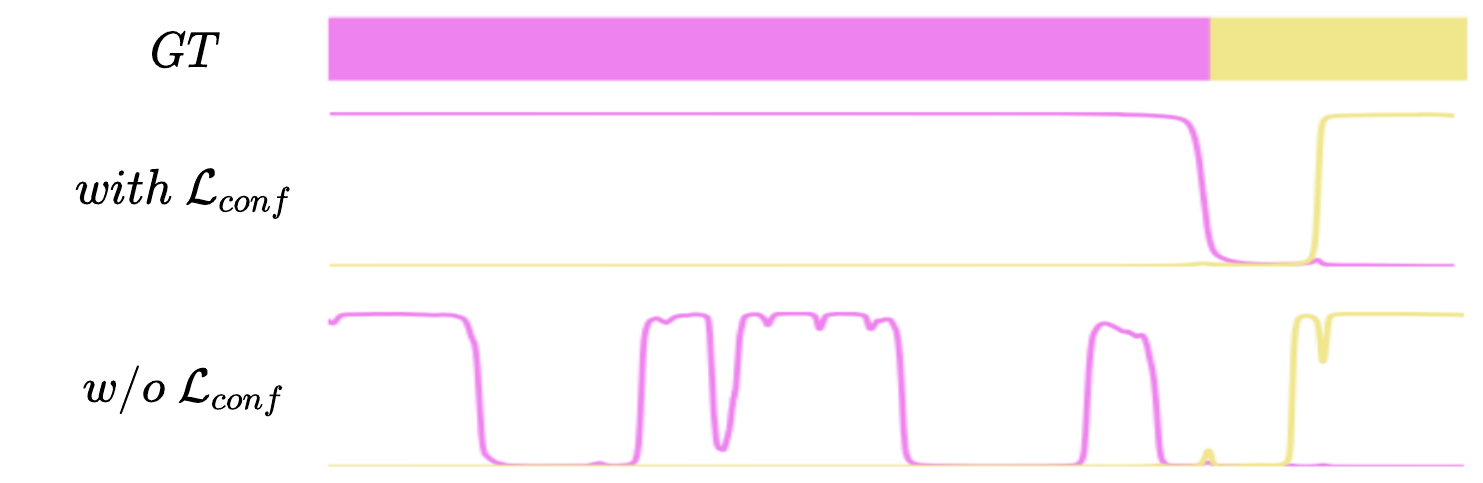}  \\
         (a) \\
         \includegraphics[trim={0.8cm 0cm 0cm 0cm},clip,width=.7\columnwidth]{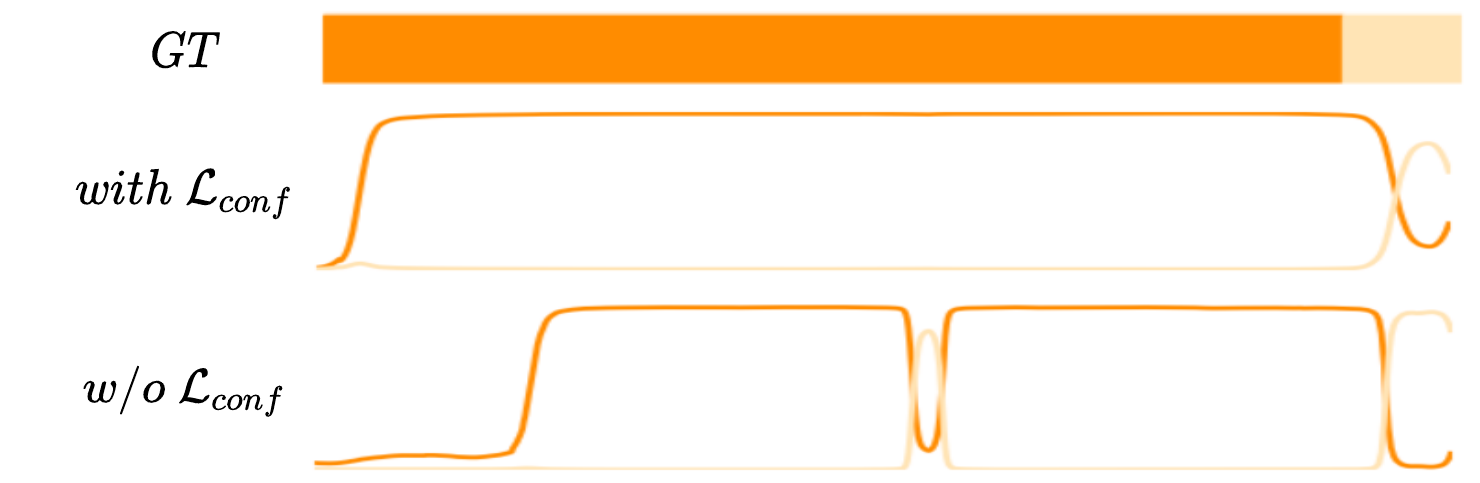} \\
         (b)
   \end{tabular}
   \caption{Impact of the confidence loss. Forcing monotonicity encourages the model to have a high confidence for all frames within an action segment (a). It also suppresses outlier frames with high confidence (b).}
   \label{fig:effect_of_conf_loss}
   \vspace{-3mm}
\end{figure}

While the smoothing loss forces a smooth transition between consecutive frames, it does not 
take the annotations into account. On the contrary, the confidence loss forces the predicted 
probabilities to monotonically decrease as the distance to the timestamps increases. This 
encourages the model to have a high confidence for all frames within an action segment, and yet 
it suppresses outlier frames that are far from the annotations and not supported by regions with 
high confidence as illustrated in Fig.~\ref{fig:effect_of_conf_loss}.

To balance the contribution of the different losses, we set the weight of the smoothing loss to 
$0.15$ as in~\cite{farha2019ms}, and the weight of the confidence loss $\beta = 0.075$. In 
Table~\ref{tab:beta}, we study the impact of $\beta$ on the performance on the 50Salads dataset. 
As shown in the table, good results are achieved for $\beta$ between $0.05$ and $0.1$.

\begin{table}[tb]\setlength{\tabcolsep}{8pt}
   \centering
   \resizebox{.9\columnwidth}{!}{%
      \begin{tabular}{lccccc}
         \toprule
          & \multicolumn{3}{c}{F1@\{10, 25, 50\}} & Edit & Acc  \\
         \midrule
         \emph{\textbf{50Salads}} \\
         \midrule
         $\mathcal{L}_{cls}$ & 65.7 & 62.6 & 50.7 & 57.7 & 72.8 \\
         $\mathcal{L}_{cls} + \alpha \mathcal{L}_{T-MSE}$  & 70.1 & 66.8 & 55.3 & 62.6 & 74.6 \\
         $\mathcal{L}_{cls} + \beta \mathcal{L}_{conf}$ & 73.2 & 70.6 & 60.1 & 65.2 & 75.3 \\
         $\mathcal{L}_{cls} + \alpha \mathcal{L}_{T-MSE} + \beta \mathcal{L}_{conf}$ & \textbf{73.9} & \textbf{70.9} & \textbf{60.1} & \textbf{66.8} & \textbf{75.6} \\
         \midrule
         \emph{\textbf{Breakfast}} \\
         \midrule
         $\mathcal{L}_{cls}$ & 60.3 & 52.8 & 36.7 & 64.2 & 60.2 \\
         $\mathcal{L}_{cls} + \alpha \mathcal{L}_{T-MSE}$  & 67.5 & 60.1 & 44.3 & 68.9 & 63.7 \\
         $\mathcal{L}_{cls} + \beta \mathcal{L}_{conf}$ & 67.6 & 60.4 & 43.7 & 68.0 & 61.6 \\
         $\mathcal{L}_{cls} + \alpha \mathcal{L}_{T-MSE} + \beta \mathcal{L}_{conf}$ & \textbf{70.5} & \textbf{63.6} & \textbf{47.4} & \textbf{69.9} & \textbf{64.1} \\
         \bottomrule
      \end{tabular}
    }
   \caption{Contribution of the different loss functions on the 50Salads and Breakfast datasets.}
   \label{tab:losses}
\end{table}


\begin{table}[tb]\setlength{\tabcolsep}{10pt}
   \centering
   \resizebox{.85\columnwidth}{!}{%
      \begin{tabular}{@{\hskip .2in}l@{\hskip 0.5in}ccc@{\hskip .2in}c@{\hskip .2in}c}
         \toprule
          & \multicolumn{3}{c}{F1@\{10, 25, 50\}} & Edit & Acc  \\
         \midrule
         $\beta = 0$  & 70.1 & 66.8 & 55.3 & 62.6 & 74.6 \\
         $\beta = 0.025$ & 70.9 & 68.8 & 57.4 & 63.4 & \textbf{76.2} \\
         $\beta = 0.05$ & 73.1 & 70.2 & 58.7 & 65.4 & 75.6 \\
         $\beta = 0.075$ & \textbf{73.9} & \textbf{70.9} & \textbf{60.1} & \textbf{66.8} & 75.6 \\
         $\beta = 0.1$ & 73.2 & 70.6 & \textbf{60.1} & 66.1 & 74.6 \\
         \bottomrule
      \end{tabular}
    }
   \caption{Impact of $\beta$ on the 50Salads dataset.}
   \vspace{-4mm}
   \label{tab:beta}
\end{table}


\subsection{Impact of the Energy Function for Action Change Detection}

Our approach generates target labels by estimating where the action labels change using the 
forward-backward estimate as in (\ref{eqn:forward_backward}). To analyze the impact of 
this estimate, we train another model that directly uses the stamp-to-stamp estimate 
(Stamp-to-Stamp (Features)) as in (\ref{eqn:s2s_energy}). As shown in Table~\ref{tab:forward_backward}, our 
approach performs better. We also tried another variant of the stamp-to-stamp energy 
function that maximizes the average probabilities of the action segments (Stamp-to-Stamp (Prob.)) instead of minimizing 
the distances to cluster centers. However, the performance is worse than the proposed energy function.


\begin{table}[tb]\setlength{\tabcolsep}{8pt}
   \centering
   \resizebox{.9\columnwidth}{!}{%
      \begin{tabular}{lccccc}
         \toprule
          & \multicolumn{3}{c}{F1@\{10, 25, 50\}} & Edit & Acc  \\
         \midrule
         \emph{\textbf{50Salads}} \\
         \midrule
         Stamp-to-Stamp (Prob.) & 67.5 & 61.8 & 48.6 & 61.1 & 68.9   \\
         Stamp-to-Stamp (Features) & 73.4 & 70.5 & 59.9 & 66.7 & 74.2  \\
         Ours &\textbf{73.9} & \textbf{70.9} & \textbf{60.1} & \textbf{66.8} & \textbf{75.6}  \\
         \midrule
         \emph{\textbf{Breakfast}} \\
         \midrule
         Stamp-to-Stamp (Prob.) &  65.7& 55.9 & 35.9 & 68.0 & 58.8  \\
         Stamp-to-Stamp (Features) & 66.3 & 59.6 & 44.4 & 67.9 & 60.1  \\
         Ours & \textbf{70.5} & \textbf{63.6} & \textbf{47.4} & \textbf{69.9} & \textbf{64.1} \\
         \bottomrule
      \end{tabular}
    }
   \caption{Impact of the energy function for action change detection on the 50Salads and Breakfast datasets.}
   \label{tab:forward_backward}
\end{table}


\subsection{Impact of the Segmentation Model $\mathcal{M}$}

In all experiments, we used a multi-stage temporal convolutional architecture based on~\cite{farha2019ms} 
and~\cite{vats2020event}. In this section we study the impact of the segmentation model on the 
performance. To this end, we apply the proposed training scheme on the original MS-TCN~\cite{farha2019ms} 
and the recently introduced MS-TCN++~\cite{li2020ms}. As shown in Table~\ref{tab:impact_of_seg_model}, 
our approach is agnostic to the segmentation model and performs well with all these models.

\begin{table}[tb]
   \centering
   \resizebox{.9\columnwidth}{!}{%
      \begin{tabular}{clccccc}
         \toprule
         Dataset & Seg. Model $\mathcal{M}$  &  \multicolumn{3}{c}{F1@\{10, 25, 50\}} & Edit & Acc  \\
         \midrule
         50Salads & MS-TCN~\cite{farha2019ms} & 71.7 & 68.7 & 57.0 & 64.0 & 74.7  \\
                  & MS-TCN++~\cite{li2020ms}  & 75.0 & 71.1 & 55.8 & 67.2 & 72.9  \\
                  & Ours & 73.9 & 70.9 & 60.1 & 66.8 & 75.6  \\
         
         \midrule
         GTEA & MS-TCN~\cite{farha2019ms} & 79.8 & 73.3 & 47.7 & 76.3 & 59.3\\
              & MS-TCN++~\cite{li2020ms} & 78.3 & 72.2 & 49.1 & 74.5 & 62.2  \\
              & Ours &  78.9 & 73.0 & 55.4 & 72.3 & 66.4 \\
         \bottomrule
      \end{tabular}
    }
   \caption{Impact of the segmentation model $\mathcal{M}$ on the 50Salads and GTEA datasets.}
   \vspace{-3mm}
   \label{tab:impact_of_seg_model}
\end{table}


\subsection{Comparison with the State-of-the-Art}
\label{sec:comparison_with_sota}

In this section, we compare our approach with recent state-of-the-art approaches for 
timestamp supervision. To the best of our knowledge, timestamp supervision has not been 
studied for the temporal action segmentation task. We, therefore, compare with similar methods in the context of action recognition~\cite{moltisanti2019action} and action localization~\cite{ma2020sf}.

Since the approach of~\cite{moltisanti2019action} assumes the testing videos are trimmed and 
does not work for long untrimmed videos, we replaced their backbone network with our segmentation 
model for a fair comparison. To this end, we initialized the plateau functions around the timestamps 
annotations of the training videos and iteratively update their parameters based on the segmentation 
model output as in~\cite{moltisanti2019action}. Results for the 50Salads, Breakfast, and GTEA datasets 
are shown in Tables~\ref{tab:conparison50salads}-\ref{tab:conparisonGTEA}, respectively. Our 
approach outperforms~\cite{moltisanti2019action} on all datasets with a large margin of 
up to $13.5\%$ frame-wise accuracy and $11.8\%$ for the F1 score with $50\%$ overlapping threshold 
on the GTEA dataset. 

We also compare our approach for timestamp supervision with other levels of supervision for the 
temporal action segmentation task. As shown in Tables~\ref{tab:conparison50salads}-\ref{tab:conparisonGTEA}, 
timestamp supervision outperforms weaker levels of supervision in the form of transcripts or sets with 
a large margin. Our approach provides a good compromise between annotation effort and performance, and 
further reduces the gap to fully supervised approaches.

\begin{table}[tb]
   \centering
   \resizebox{.9\columnwidth}{!}{%
      \begin{tabular}{clccccc}
         \toprule
         Supervision & Method & \multicolumn{3}{c}{F1@\{10, 25, 50\}} & Edit & Acc  \\
         \midrule
         Full & 
         MS-TCN~\cite{farha2019ms} & 76.3 & 74.0 & 64.5 & 67.9 & 80.7  \\
         & MS-TCN++~\cite{li2020ms} & 80.7 & 78.5 & 70.1 & 74.3 & 83.7  \\
         & BCN~\cite{wang2020boundary} & 82.3 & 81.3 & 74.0 & 74.3 & 84.4  \\
         & ASRF~\cite{ishikawa2020alleviating} & 84.9 & 83.5 & 77.3 & 79.3 & 84.5  \\
         \midrule
         Timestamps &  
         \footnotesize{Seg. Model $\mathcal{M}$ + plateau~\cite{moltisanti2019action}} & 71.2 & 68.2 & 56.1 & 62.6 & 73.9 \\
         & Ours &  \textbf{73.9} & \textbf{70.9} & \textbf{60.1} & \textbf{66.8} & \textbf{75.6}   \\
         \midrule
         Transcripts &  
         CDFL~\cite{li2019weakly}    & - & - & - & - & 54.7 \\
         & NN-Viterbi~\cite{richard2018neuralnetwork} & - & - & - & - & 49.4  \\
         & HMM-RNN~\cite{richard2017weakly}  & - & - & - & - & 45.5  \\
         \bottomrule
      \end{tabular}
    }
   \caption{Comparison with different levels of supervision on the 50Salads dataset.}
   \label{tab:conparison50salads}
   \vspace{-0.5mm}
\end{table}


\begin{table}[tb]
   \centering
   \resizebox{.9\columnwidth}{!}{%
      \begin{tabular}{clccccc}
         \toprule
         Supervision & Method & \multicolumn{3}{c}{F1@\{10, 25, 50\}} & Edit & Acc  \\
         \midrule
         Full & 
         MS-TCN~\cite{farha2019ms} & 52.6 & 48.1 & 37.9 & 61.7 & 66.3  \\
         & MS-TCN++~\cite{li2020ms} & 64.1 & 58.6 & 45.9 & 65.6 & 67.6  \\
         & BCN~\cite{wang2020boundary} & 68.7 & 65.5 & 55.0 & 66.2 & 70.4  \\
         & ASRF~\cite{ishikawa2020alleviating} & 74.3 & 68.9 & 56.1 & 72.4 & 67.6  \\
         \midrule
         Timestamps & 
         \footnotesize{Seg. Model $\mathcal{M}$ + plateau~\cite{moltisanti2019action}} &     65.5 & 59.1 & 43.2 & 65.9 & 63.5 \\
         & Ours &  \textbf{70.5} & \textbf{63.6} & \textbf{47.4} & \textbf{69.9} & \textbf{64.1}  \\
         \midrule
         Transcripts &
         CDFL~\cite{li2019weakly}    & - & - & - & - & 50.2 \\
         & MuCon~\cite{souri2019fast} & - & - & - & - & 47.1  \\
         & D$^{3}$TW~\cite{chang2019d3tw} & - & - & - & - & 45.7  \\
         & NN-Viterbi~\cite{richard2018neuralnetwork} & - & - & - & - & 43.0  \\
         & TCFPN~\cite{ding2018weakly}  & - & - & - & - & 38.4  \\
         & HMM-RNN~\cite{richard2017weakly}  & - & - & - & - & 33.3  \\
         & ECTC~\cite{huang2016connectionist}  & - & - & - & - & 27.7  \\
         \midrule
         Sets & 
         SCT~\cite{fayyaz2020set} & - & - & - & - & 30.4  \\
         & SCV~\cite{li2020set} & - & - & - & - & 30.2  \\
         & Action Sets~\cite{richard2018action} & - & - & - & - & 23.3 \\
         \bottomrule
      \end{tabular}
    }
   \caption{Comparison with different levels of supervision on the Breakfast dataset.}
   \label{tab:conparisonbreakfast}
   \vspace{-0.5mm}
\end{table}


\begin{table}[tb]
   \centering
   \resizebox{.9\columnwidth}{!}{%
      \begin{tabular}{clccccc}
         \toprule
         Supervision & Method & \multicolumn{3}{c}{F1@\{10, 25, 50\}} & Edit & Acc  \\
         \midrule
         Full & 
         MS-TCN~\cite{farha2019ms} & 85.8 & 83.4 & 69.8 & 79.0 & 76.3  \\
         & MS-TCN++~\cite{li2020ms} & 88.8 & 85.7 & 76.0 & 83.5 & 80.1  \\
         & BCN~\cite{wang2020boundary} & 88.5 & 87.1 & 77.3 & 84.4 & 79.8  \\
         & ASRF~\cite{ishikawa2020alleviating} & 89.4 & 87.8 & 79.8 & 83.7 & 77.3  \\
         \midrule
         Timestamps &  
         \footnotesize{Seg. Model $\mathcal{M}$ + plateau~\cite{moltisanti2019action}} & 74.8 & 68.0 & 43.6 & \textbf{72.3} & 52.9   \\
         & Ours &  \textbf{78.9} & \textbf{73.0} & \textbf{55.4} & \textbf{72.3} & \textbf{66.4}\\
         \bottomrule
      \end{tabular}
    }
   \caption{Comparison with different levels of supervision on the GTEA dataset.}
   \label{tab:conparisonGTEA}
   \vspace{-3mm}
\end{table}


Timestamp supervision has recently been studied for action localization in~\cite{ma2020sf}. In their 
approach, they use the model confidence to sample foreground action frames and background frames for 
training. To compare with~\cite{ma2020sf}, we use the same setup and the provided human annotations to train our model and report mean 
average precision at different overlapping thresholds. Table~\ref{tab:sfnetmap} shows the results on 
the GTEA and BEOID~\cite{damen2014you} datasets. Our approach outperforms Ma~\etal~\cite{ma2020sf} with 
a large margin of $5.4\%$ average mAP on GTEA and $4.3\%$ on the BEOID dataset. In contrast to~\cite{ma2020sf} 
where only the frames with high confidence are used for training, our approach detects action 
changes and generates a target label for each frame in the training videos.

Finally, we compare our approach with the semi-supervised setup on the Breakfast dataset 
proposed in Kuehne~\etal~\cite{kuehne2020hybrid}. In this setup, the training videos are annotated with 
the transcript of actions and a fraction of the frames as well. Compared to the timestamp 
supervision, this setup provides annotations for much more frames. 
Since the timestamps are randomly sampled from the video, there are sometimes multiple timestamps for one action and not all actions are annotated as reported in the supplementary material. 
As shown in Table~\ref{tab:semi_sup}, 
our approach outperforms~\cite{kuehne2020hybrid} with a large margin. While the 
approach of~\cite{kuehne2020hybrid} relies on an expensive Viterbi decoding during inference, our 
approach directly predicts the frame-wise labels.


\begin{table}[tb]\setlength{\tabcolsep}{8pt}
   \centering
   \resizebox{.77\columnwidth}{!}{%
      \begin{tabular}{@{\hskip .3in}l@{\hskip .5in}ccccc}
         \toprule
         mAP@IoU & 0.1 & 0.3 & 0.5 & 0.7 & Avg  \\
         \midrule
         \emph{\textbf{GTEA}} \\
         \midrule
         SF-Net~\cite{ma2020sf} & 58.0 & 37.9 & 19.3 & 11.9& 31.0 \\
         Ours & \textbf{60.2} & \textbf{44.7} & \textbf{28.8} & \textbf{12.2} & \textbf{36.4} \\
         \midrule
         \emph{\textbf{BEOID}} \\
         \midrule
         SF-Net~\cite{ma2020sf} & 62.9 & \textbf{40.6} & 16.7 & 3.5 & 30.1 \\
         Ours & \textbf{71.5} & 40.3 & \textbf{20.3} & \textbf{5.5} & \textbf{34.4} \\
         \bottomrule
      \end{tabular}
    }
   \caption{Comparison with SF-Net~\cite{ma2020sf} for action localization with timestamp supervision on the GTEA and BEOID datasets.}
   \label{tab:sfnetmap}
\end{table}


\begin{table}[tb]\setlength{\tabcolsep}{24pt}
   \centering
   \resizebox{.78\columnwidth}{!}{%
      \begin{tabular}{clc}
         \toprule
         Fraction & Method  & Acc  \\
         \midrule
         0.1 & HMM-RNN~\cite{kuehne2020hybrid} & 60.9 \\
           & Ours & \textbf{68.4}  \\
         \midrule
         0.01 & HMM-RNN~\cite{kuehne2020hybrid} & 58.8 \\ 
           & Ours & \textbf{67.4}   \\
        \bottomrule
      \end{tabular}
    }
   \caption{Comparison with Kuehne~\etal~\cite{kuehne2020hybrid} on the Breakfast dataset with semi-supervised setup.}
   \label{tab:semi_sup}
   \vspace{-3mm}
\end{table}



\section{Conclusion}
In this paper, we proposed an approach to train a temporal action segmentation model using 
only timestamps annotations. Our approach combines the model predictions with the timestamps 
annotations for estimating where the action labels change. We further introduced a confidence loss that enforces monotonicity on the model confidence. The loss encourages high confidence values for all frames within an action segment and suppresses outlier frames.
Results on four 
datasets show that models trained with timestamp supervision achieve comparable performance 
to the fully supervised setup. 
The proposed approach is model agnostic and can be applied to any segmentation 
model.

\small{\noindent\textbf{\\Acknowledgements:} The work has been funded by the Deutsche Forschungsgemeinschaft (DFG, German Research Foundation) – GA 1927/4-2 (FOR 2535 Anticipating Human Behavior) and the ERC Starting Grant ARCA (677650).}

\clearpage

\section*{Supplementary Material}

\def\thesection{\Alph{section}}
\renewcommand{\thetable}{\Alph{table}}
\setcounter{section}{0}
\setcounter{table}{0}

\hyphenation{time-stamp}
\hyphenation{time-stamps}







We evaluate our model on additional settings for generating the timestamps annotations. 
We further analyze the impact of noise on the performance.

\section{Frame Selection for the Timestamp Annotations}

In the paper, we randomly select one frame from each action segment in the training 
videos. Table~\ref{tab:ts_annotations} shows results for two additional settings. While using the center frame of each segment achieves comparable results to a random frame, the performance drops when the start frame is used. Humans, however, would not annotate the start frame since it is more ambiguous (see Fig.~4 in~\cite{ma2020sf}).

\begin{table}[ht]\setlength{\tabcolsep}{8pt}
   \centering
   \resizebox{.8\columnwidth}{!}{%
      \begin{tabular}{@{\hskip .1in}l@{\hskip .5in}ccc@{\hskip .2in}c@{\hskip .2in}c}
         \toprule
         Timestamps & \multicolumn{3}{c}{F1@\{10, 25, 50\}} & Edit & Acc  \\
         \midrule
         
         Start frame  &         65.5  &         52.2  &         28.0  &         70.4  &         51.2 \\
         Center frame & \textbf{70.8} &         63.5  &         45.4  & \textbf{71.3} &         61.3 \\
         Random       &         70.5  & \textbf{63.6} & \textbf{47.4} &         69.9  & \textbf{64.1}  \\
         \bottomrule
      \end{tabular}
    }
   \caption{Using start, center, or random frame as timestamp for each action on the Breakfast dataset.}
   \label{tab:ts_annotations}
\end{table}


\section{Human Annotations vs Generated Annotations}

In Table~\ref{tab:gtea}, we directly compare the human annotations from~\cite{ma2020sf} with 
simulated annotations on the GTEA dataset. The results show that the performance with random 
sampling is very close to real annotations.

\begin{table}[ht]\setlength{\tabcolsep}{8pt}
   \centering
   \resizebox{.8\columnwidth}{!}{%
      \begin{tabular}{@{\hskip .1in}l@{\hskip .3in}ccccc}
         \toprule
         mAP@IoU & 0.1 & 0.3 & 0.5 & 0.7 & Avg  \\
         \midrule
         Human annotations & \textbf{60.2} & 44.7 & \textbf{28.8} & \textbf{12.2} & \textbf{36.4} \\
         Random sampling & 59.7  &  \textbf{46.3}  & 26.0  & 10.4  &  35.6 \\
         \bottomrule
      \end{tabular}
    }
   \caption{Human vs.\ generated annotations on the GTEA dataset.}
   \label{tab:gtea}
\end{table}


\section{Impact of Noise}

In Table~10 in the paper, the timestamps are randomly sampled from the videos. Thus, there are sometimes multiple timestamps for one action and not all actions are annotated. Table~\ref{tab:semi_sup} shows the percentage of action segments with 0, 1, or $>$1 timestamps (TS). 
As shown in the table, our approach is robust to annotation errors.

\begin{table}[ht]
   \centering
   \resizebox{.9\columnwidth}{!}{%
      \begin{tabular}{c|ccc|c}
         \toprule
         Fraction & \% actions with 0 TS & \% with 1 TS & \% with $>$ 1 TS & Acc  \\
         \midrule
         0.1     &   3.5  &   3.4  &  93.1  & 68.4  \\
         0.01    &  27.5  &  21.6  &  50.9  & 67.4   \\
         0.0032  &  49.0  &  25.2  &  25.8  & 61.7 \\
         0.0032  &  0  &  100  &  0  & 64.1 \\
        \bottomrule
      \end{tabular}
    }
   \caption{Percentage of actions with 0, 1, or $>$1 timestamps (TS) using the protocol of Table~10. The last row is the protocol of Table~7. 
   }
   \label{tab:semi_sup}
\end{table}





\clearpage

{\small
\bibliographystyle{ieee_fullname}
\bibliography{egbib}

\begin{thebibliography}{10}\itemsep=-1pt

\bibitem{farha2019ms}
Yazan Abu~Farha and Juergen Gall.
\newblock {MS-TCN}: {M}ulti-stage temporal convolutional network for action
  segmentation.
\newblock In {\em IEEE Conference on Computer Vision and Pattern Recognition
  (CVPR)}, pages 3575--3584, 2019.

\bibitem{bearman2016s}
Amy Bearman, Olga Russakovsky, Vittorio Ferrari, and Li Fei-Fei.
\newblock What’s the point: Semantic segmentation with point supervision.
\newblock In {\em European Conference on Computer Vision (ECCV)}, pages
  549--565. Springer, 2016.

\bibitem{bojanowski2014weakly}
Piotr Bojanowski, R{\'e}mi Lajugie, Francis Bach, Ivan Laptev, Jean Ponce,
  Cordelia Schmid, and Josef Sivic.
\newblock Weakly supervised action labeling in videos under ordering
  constraints.
\newblock In {\em European Conference on Computer Vision (ECCV)}, pages
  628--643. Springer, 2014.

\bibitem{carreira2017quo}
Joao Carreira and Andrew Zisserman.
\newblock Quo vadis, action recognition? {A} new model and the kinetics
  dataset.
\newblock In {\em IEEE Conference on Computer Vision and Pattern Recognition
  (CVPR)}, pages 6299--6308, 2017.

\bibitem{chang2019d3tw}
Chien-Yi Chang, De-An Huang, Yanan Sui, Li Fei-Fei, and Juan~Carlos Niebles.
\newblock {D$^{3}$TW}: {D}iscriminative differentiable dynamic time warping for
  weakly supervised action alignment and segmentation.
\newblock In {\em IEEE Conference on Computer Vision and Pattern Recognition
  (CVPR)}, pages 3546--3555, 2019.

\bibitem{cheron2018flexible}
Guilhem Ch{\'e}ron, Jean-Baptiste Alayrac, Ivan Laptev, and Cordelia Schmid.
\newblock A flexible model for training action localization with varying levels
  of supervision.
\newblock In {\em Advances in Neural Information Processing Systems (NeurIPS)},
  pages 942--953, 2018.

\bibitem{damen2014you}
Dima Damen, Teesid Leelasawassuk, Osian Haines, Andrew Calway, and Walterio
  Mayol-Cuevas.
\newblock You-do, {I}-learn: Discovering task relevant objects and their modes
  of interaction from multi-user egocentric video.
\newblock In {\em British Machine Vision Conference (BMVC)}, 2014.

\bibitem{ding2018weakly}
Li Ding and Chenliang Xu.
\newblock Weakly-supervised action segmentation with iterative soft boundary
  assignment.
\newblock In {\em IEEE Conference on Computer Vision and Pattern Recognition
  (CVPR)}, pages 6508--6516, 2018.

\bibitem{duchenne2009automatic}
Olivier Duchenne, Ivan Laptev, Josef Sivic, Francis Bach, and Jean Ponce.
\newblock Automatic annotation of human actions in video.
\newblock In {\em IEEE International Conference on Computer Vision (ICCV)},
  pages 1491--1498, 2009.

\bibitem{fathi2011learning}
Alireza Fathi, Xiaofeng Ren, and James~M Rehg.
\newblock Learning to recognize objects in egocentric activities.
\newblock In {\em IEEE Conference on Computer Vision and Pattern Recognition
  (CVPR)}, pages 3281--3288, 2011.

\bibitem{fayyaz2020set}
Mohsen Fayyaz and Jurgen Gall.
\newblock {SCT}: {S}et constrained temporal transformer for set supervised
  action segmentation.
\newblock In {\em IEEE Conference on Computer Vision and Pattern Recognition
  (CVPR)}, 2020.

\bibitem{feichtenhofer2019slowfast}
Christoph Feichtenhofer, Haoqi Fan, Jitendra Malik, and Kaiming He.
\newblock Slowfast networks for video recognition.
\newblock In {\em IEEE International Conference on Computer Vision (ICCV)},
  2019.

\bibitem{huang2016connectionist}
De-An Huang, Li Fei-Fei, and Juan~Carlos Niebles.
\newblock Connectionist temporal modeling for weakly supervised action
  labeling.
\newblock In {\em European Conference on Computer Vision (ECCV)}, pages
  137--153. Springer, 2016.

\bibitem{huang2020improving}
Yifei Huang, Yusuke Sugano, and Yoichi Sato.
\newblock Improving action segmentation via graph-based temporal reasoning.
\newblock In {\em IEEE Conference on Computer Vision and Pattern Recognition
  (CVPR)}, pages 14024--14034, 2020.

\bibitem{ishikawa2020alleviating}
Yuchi Ishikawa, Seito Kasai, Yoshimitsu Aoki, and Hirokatsu Kataoka.
\newblock Alleviating over-segmentation errors by detecting action boundaries.
\newblock In {\em IEEE Winter Conference on Applications of Computer Vision
  (WACV)}, pages 2322--2331, 2021.

\bibitem{kuehne2014language}
Hilde Kuehne, Ali Arslan, and Thomas Serre.
\newblock The language of actions: Recovering the syntax and semantics of
  goal-directed human activities.
\newblock In {\em IEEE Conference on Computer Vision and Pattern Recognition
  (CVPR)}, pages 780--787, 2014.

\bibitem{kuehne2016end}
Hilde Kuehne, Juergen Gall, and Thomas Serre.
\newblock An end-to-end generative framework for video segmentation and
  recognition.
\newblock In {\em IEEE Winter Conference on Applications of Computer Vision
  (WACV)}, 2016.

\bibitem{kuehne2017weakly}
Hilde Kuehne, Alexander Richard, and Juergen Gall.
\newblock Weakly supervised learning of actions from transcripts.
\newblock {\em Computer Vision and Image Understanding}, 163:78--89, 2017.

\bibitem{kuehne2020hybrid}
Hilde Kuehne, Alexander Richard, and Juergen Gall.
\newblock A {H}ybrid {RNN}-{HMM} approach for weakly supervised temporal action
  segmentation.
\newblock {\em IEEE Transactions on Pattern Analysis and Machine Intelligence
  (TPAMI)}, 42(04):765--779, 2020.

\bibitem{lea2017temporal}
Colin Lea, Michael~D. Flynn, Rene Vidal, Austin Reiter, and Gregory~D. Hager.
\newblock Temporal convolutional networks for action segmentation and
  detection.
\newblock In {\em IEEE Conference on Computer Vision and Pattern Recognition
  (CVPR)}, 2017.

\bibitem{lea2016segmental}
Colin Lea, Austin Reiter, Ren{\'e} Vidal, and Gregory~D Hager.
\newblock Segmental spatiotemporal {CNN}s for fine-grained action segmentation.
\newblock In {\em European Conference on Computer Vision (ECCV)}, pages 36--52.
  Springer, 2016.

\bibitem{lee2020background}
Pilhyeon Lee, Youngjung Uh, and Hyeran Byun.
\newblock Background suppression network for weakly-supervised temporal action
  localization.
\newblock In {\em AAAI Conference on Artificial Intelligence}, pages
  11320--11327, 2020.

\bibitem{lee2021weakly}
Pilhyeon Lee, Jinglu Wang, Yan Lu, and Hyeran Byun.
\newblock Weakly-supervised temporal action localization by uncertainty
  modeling.
\newblock In {\em AAAI Conference on Artificial Intelligence}, 2021.

\bibitem{lei2018temporal}
Peng Lei and Sinisa Todorovic.
\newblock Temporal deformable residual networks for action segmentation in
  videos.
\newblock In {\em IEEE Conference on Computer Vision and Pattern Recognition
  (CVPR)}, pages 6742--6751, 2018.

\bibitem{li2019weakly}
Jun Li, Peng Lei, and Sinisa Todorovic.
\newblock Weakly supervised energy-based learning for action segmentation.
\newblock In {\em IEEE International Conference on Computer Vision (ICCV)},
  pages 6243--6251, 2019.

\bibitem{li2020set}
Jun Li and Sinisa Todorovic.
\newblock Set-constrained viterbi for set-supervised action segmentation.
\newblock In {\em IEEE Conference on Computer Vision and Pattern Recognition
  (CVPR)}, pages 10820--10829, 2020.

\bibitem{li2020ms}
Shijie Li, Yazan Abu~Farha, Yun Liu, Ming-Ming Cheng, and Juergen Gall.
\newblock {MS-TCN}++: {M}ulti-stage temporal convolutional network for action
  segmentation.
\newblock {\em IEEE Transactions on Pattern Analysis and Machine Intelligence
  (TPAMI)}, 2020.

\bibitem{ma2020sf}
Fan Ma, Linchao Zhu, Yi Yang, Shengxin Zha, Gourab Kundu, Matt Feiszli, and
  Zheng Shou.
\newblock {SF}-{N}et: {S}ingle-frame supervision for temporal action
  localization.
\newblock In {\em European Conference on Computer Vision (ECCV)}, 2020.

\bibitem{mac2019learning}
Khoi-Nguyen~C. Mac, Dhiraj Joshi, Raymond~A. Yeh, Jinjun Xiong, Rogerio~S.
  Feris, and Minh~N. Do.
\newblock Learning motion in feature space: Locally-consistent deformable
  convolution networks for fine-grained action detection.
\newblock In {\em IEEE International Conference on Computer Vision (ICCV)},
  2019.

\bibitem{mettes2016spot}
Pascal Mettes, Jan~C Van~Gemert, and Cees~GM Snoek.
\newblock Spot on: Action localization from pointly-supervised proposals.
\newblock In {\em European Conference on Computer Vision (ECCV)}, pages
  437--453. Springer, 2016.

\bibitem{moltisanti2019action}
Davide Moltisanti, Sanja Fidler, and Dima Damen.
\newblock Action recognition from single timestamp supervision in untrimmed
  videos.
\newblock In {\em IEEE Conference on Computer Vision and Pattern Recognition
  (CVPR)}, pages 9915--9924, 2019.

\bibitem{pirsiavash2014parsing}
Hamed Pirsiavash and Deva Ramanan.
\newblock Parsing videos of actions with segmental grammars.
\newblock In {\em IEEE Conference on Computer Vision and Pattern Recognition
  (CVPR)}, pages 612--619, 2014.

\bibitem{richard2017weakly}
Alexander Richard, Hilde Kuehne, and Juergen Gall.
\newblock Weakly supervised action learning with {RNN} based fine-to-coarse
  modeling.
\newblock In {\em IEEE Conference on Computer Vision and Pattern Recognition
  (CVPR)}, pages 754--763, 2017.

\bibitem{richard2018action}
Alexander Richard, Hilde Kuehne, and Juergen Gall.
\newblock Action sets: {W}eakly supervised action segmentation without ordering
  constraints.
\newblock In {\em IEEE Conference on Computer Vision and Pattern Recognition
  (CVPR)}, pages 5987--5996, 2018.

\bibitem{richard2018neuralnetwork}
Alexander Richard, Hilde Kuehne, Ahsan Iqbal, and Juergen Gall.
\newblock Neural{N}etwork-{V}iterbi: A framework for weakly supervised video
  learning.
\newblock In {\em IEEE Conference on Computer Vision and Pattern Recognition
  (CVPR)}, pages 7386--7395, 2018.

\bibitem{simonyan2014two}
Karen Simonyan and Andrew Zisserman.
\newblock Two-stream convolutional networks for action recognition in videos.
\newblock In {\em Advances in Neural Information Processing Systems (NeurIPS)},
  pages 568--576, 2014.

\bibitem{souri2019fast}
Yaser Souri, Mohsen Fayyaz, Luca Minciullo, Gianpiero Francesca, and Juergen
  Gall.
\newblock Fast weakly supervised action segmentation using mutual consistency.
\newblock {\em arXiv preprint arXiv:1904.03116}, 2019.

\bibitem{stein2013combining}
Sebastian Stein and Stephen~J McKenna.
\newblock Combining embedded accelerometers with computer vision for
  recognizing food preparation activities.
\newblock In {\em ACM International Joint Conference on Pervasive and
  Ubiquitous Computing}, pages 729--738, 2013.

\bibitem{vats2020event}
Kanav Vats, Mehrnaz Fani, Pascale Walters, David~A Clausi, and John Zelek.
\newblock Event detection in coarsely annotated sports videos via parallel
  multi-receptive field {1D} convolutions.
\newblock In {\em IEEE Conference on Computer Vision and Pattern Recognition
  Workshops (CVPRW)}, pages 882--883, 2020.

\bibitem{vo2014stochastic}
Nam~N Vo and Aaron~F Bobick.
\newblock From stochastic grammar to bayes network: Probabilistic parsing of
  complex activity.
\newblock In {\em IEEE Conference on Computer Vision and Pattern Recognition
  (CVPR)}, pages 2641--2648, 2014.

\bibitem{wang2020boundary}
Zhenzhi Wang, Ziteng Gao, Limin Wang, Zhifeng Li, and Gangshan Wu.
\newblock Boundary-aware cascade networks for temporal action segmentation.
\newblock In {\em European Conference on Computer Vision (ECCV)}, 2020.

\end{thebibliography}
}


\end{document}